\documentclass[final]{cvpr}
\usepackage{times}
\usepackage{epsfig}
\usepackage{graphicx}
\usepackage{amsmath}
\usepackage{amssymb}
\usepackage[utf8]{inputenc} 
\usepackage[T1]{fontenc}    
\usepackage{url}            
\usepackage{booktabs}       
\usepackage{amsfonts}       
\usepackage{nicefrac}       
\usepackage{microtype}      
\usepackage{subfiles}
\usepackage{cite}
\usepackage{bm}
\usepackage{color}
\usepackage{flexisym}
            
\usepackage{pifont}
\newcommand{\cmark}{\text{\ding{51}}}
\newcommand{\xmark}{\text{\ding{55}}}

\usepackage{marvosym}
\newcommand\blfootnote[1]{%
  \begingroup
  \renewcommand\thefootnote{}\footnote{#1}%
  \addtocounter{footnote}{-1}%
  \endgroup
}

\usepackage{array}
\usepackage{multirow}
\newcolumntype{S}{>{\centering\arraybackslash}m{0.9cm}}
\newcolumntype{M}{>{\centering\arraybackslash}m{1.2cm}}
\newcolumntype{L}{>{\centering\arraybackslash}m{1.4cm}}
\usepackage[table]{xcolor}
\definecolor{mygray}{gray}{.95}
\definecolor{mylightergray}{gray}{.99}
\definecolor{mygreen}{RGB}{10, 179, 33}
\usepackage{caption} 
\makeatletter
\newcommand{\thickhline}{%
    \noalign {\ifnum 0=`}\fi \hrule height 1pt
    \futurelet \reserved@a \@xhline
}
\newcolumntype{"}{@{\vrule width 1pt}}
\makeatother

\usepackage[ruled, lined, linesnumbered, commentsnumbered, longend]{algorithm2e}
\SetAlFnt{\footnotesize}
\SetAlCapFnt{\footnotesize}
\SetAlCapNameFnt{\footnotesize}
\SetKwInput{KwParam}{Parameter}
\SetKwInOut{KwIn}{Input}
\SetKwInOut{KwOut}{Output}

\SetCommentSty{mycommfont}
\SetAlgoCaptionSeparator{}

\usepackage[pagebackref=true,breaklinks=true,colorlinks,bookmarks=false]{hyperref}

\def\cvprPaperID{1439} 
\def\confYear{CVPR 2021}

\begin{document}

\title{Context Modeling in 3D Human Pose Estimation: A Unified Perspective}

\author{Xiaoxuan~Ma$^{1,3*}$, \quad Jiajun~Su$^{2*}$,\quad 
Chunyu~Wang$^{4}$, \quad
Hai~Ci$^{1,5}$, \quad  
Yizhou~Wang $^{1}$ \\ 
\textsuperscript{1}Dept. of Computer Science, Center on Frontiers of Computing Studies, Peking University \\
\textsuperscript{2}Center for Data Science, Adv. Inst. of Info. Tech., Peking University \\
\textsuperscript{3}Advanced Innovation Center For Future Visual Entertainment (AICFVE), Beijing Film Academy
\\
\textsuperscript{4}Microsoft Research Asia   \quad
\textsuperscript{5}Deepwise AI Lab \\

{\tt\small 
\{maxiaoxuan, sujiajun, cihai, yizhou.wang\}@pku.edu.cn, chnuwa@microsoft.com}
}
\maketitle

\begin{abstract}
\blfootnote{* denotes equal contribution.}
Estimating $3$D human pose from a single image suffers from severe ambiguity since multiple $3$D joint configurations may have the same $2$D projection. The state-of-the-art methods often rely on context modeling methods such as pictorial structure model (PSM) or graph neural network (GNN) to reduce ambiguity. However, there is no study that rigorously compares them side by side. So we first present a general formula for context modeling in which both PSM and GNN are its special cases. By comparing the two methods, we found that the end-to-end training scheme in GNN and the limb length constraints in PSM are two complementary factors to improve results. To combine their advantages, we propose \textbf{ContextPose} based on attention mechanism that allows enforcing soft limb length constraints in a deep network. The approach effectively reduces the chance of getting absurd $3$D pose estimates with incorrect limb lengths and achieves state-of-the-art results on two benchmark datasets. More importantly, the introduction of limb length constraints into deep networks enables the approach to achieve much better generalization performance. 
\end{abstract}

\section{Introduction}
Monocular $3$D human pose estimation has attracted much attention \cite{martinez2017simple,bogo2016keep,mehta2017vnect,sun2018integral,zhou20153d,wang2014robust} because it can benefit many applications such as virtual reality and intelligent video analysis. The task is more difficult than $2$D pose estimation \cite{tompson2014joint,Sun_2019_CVPR,cao2017realtime} because it needs to estimate relative depth between body joints which suffers from severe ambiguity. Psychology experiments \cite{biederman1982scene} show that \emph{context} plays an important role in resolving ambiguity in human visual system. Following this idea, body joints can serve as mutual context to each other in human pose estimation--- localizing one facilitates the localization of the other. For example, elbow is more likely to be found at a distance from shoulder depending on the length of upper arm. Some work \cite{hassan2019resolving} also explores surrounding environment as context for joints to further narrow down the space.

\begin{figure}
	\centering
	\includegraphics[width=2.8in]{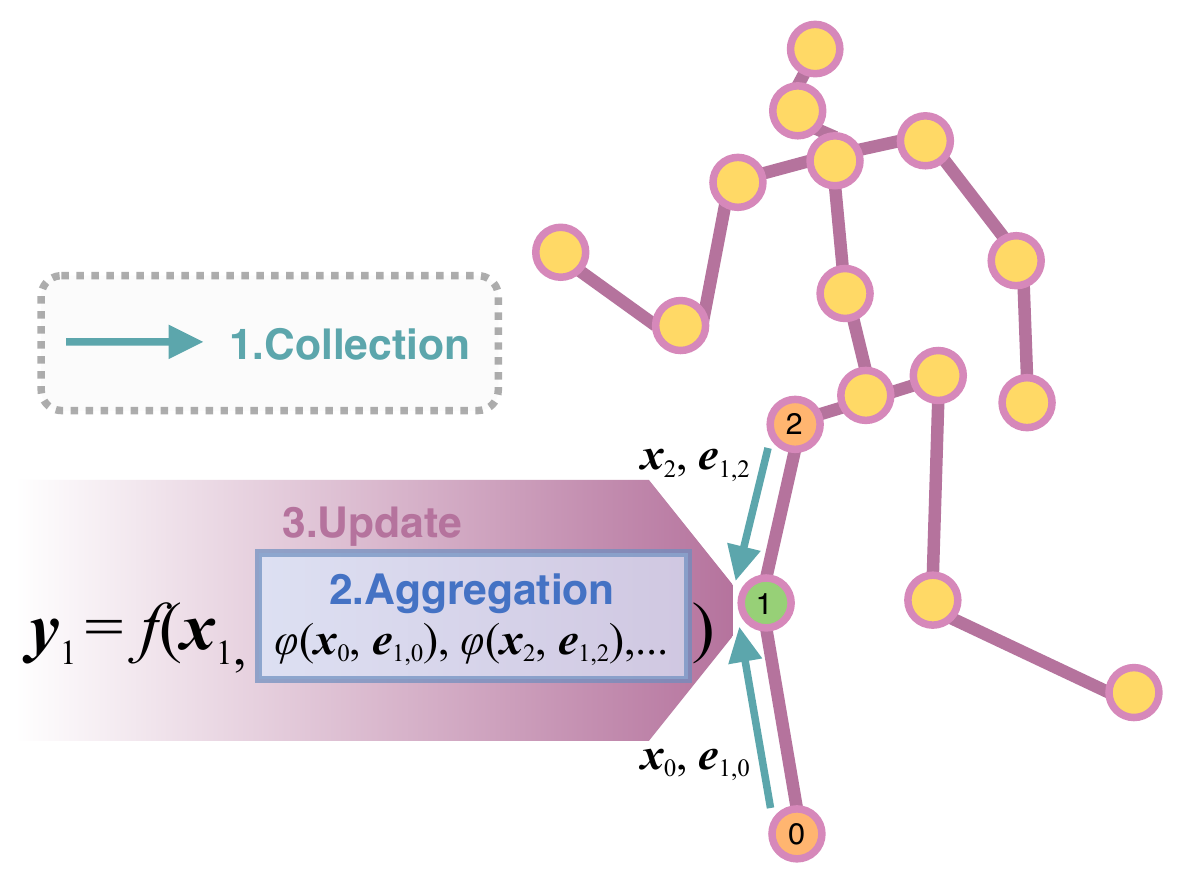}
	\caption{A general formula of context modeling in the $3$D human pose estimation task. To update features of a particular joint, the approach first collects features from its contextual joints (defined by the input graph structure), aggregates the collected features, and uses the features to update the joint of interest. }
	\label{fig:general_formula}
\end{figure}

The success of CNN in $2$D pose estimation \cite{tompson2014joint,cao2017realtime,newell2016stacked} has promoted a shift from model-based $3$D pose estimators \cite{choo2001people,lee2004human,sminchisescu2001covariance,wei2009modeling} to discriminative ones \cite{sun2018integral,martinez2017simple,Ci_2019_ICCV,kanazawa2018end}. In particular, Martinez \etal \cite{martinez2017simple} propose to estimate $3$D pose from estimated $2$D pose by a Fully Connected Network (FCN). It achieves notably smaller error than previous methods due to its strong capability of fitting large amounts of data and improved $2$D pose estimation accuracy. But it does not explicitly explore context which may result in poor results in challenging cases \cite{Ci_2019_ICCV}.

GNN \cite{defferrard2016convolutional} computes features for each node by aggregating those of its neighbors. The interaction among nodes makes it suitable for modeling context. For example, Ci \etal \cite{Ci_2019_ICCV} treat each joint as a node and perform feature passing among the nodes to estimate their $3$D locations. The method is more robust to inaccurate $2$D poses which validates the values of context.
But they cannot explicitly model spatial relation between joints such as limb length constraints which is a big limitation--- limb length is useful to reduce ambiguity when some joints are occluded.

\vspace{0.2em}

PSM \cite{amin2013multi,kostrikov2014depth,PavlakosZDD17,belagiannis20143D,qiu2019cross} had been commonly used for both $2$D and $3$D pose estimation before deep networks dominate the field. The key idea is to determine optimal joint locations by simultaneously considering their appearance and spatial relation. For example, Qiu \etal \cite{qiu2019cross} divide the $3$D motion space by regular voxels and assign each joint to the optimal voxel by minimizing an energy function defined on all joints. The approach may get accurate $3$D estimates for occluded joints based on their neighbors. Some works \cite{chen2014articulated,PavlakosZDD17,qiu2019cross} also combine PSM with deep learning by first applying CNN to estimate features and then using PSM to do inference on the features. However, the improvement is limited because it cannot be trained end-to-end.

\vspace{0.2em}

To our best knowledge, there is no work discussing the pros and cons of PSM \cite{amin2013multi,kostrikov2014depth,PavlakosZDD17,belagiannis20143D} and GNN \cite{Ci_2019_ICCV,zhao2019semantic} since they were developed in different fields. But this is actually very important. To that end, starting from their standard formulation, we develop a general formula for the two methods which allows us to clearly understand their relations and differences. In the meanwhile, we can compare their advantages and disadvantages side by side.
The basic idea is sketched in Figure \ref{fig:general_formula}. It has three steps: for each joint of interest, it first \emph{collects} features from its contextual joints which are determined by the input human graph. Then it \emph{aggregates} the collected features as context which in turn is used to \emph{update} the features of the joint.

\vspace{0.2em}

In particular, we find in our empirical study that the GNN-based methods \cite{Ci_2019_ICCV,zhao2019semantic,martinez2017simple} powered by end-to-end learning get more accurate estimates than PSM in general cases. We believe this is mainly because deep neural networks have strong capability to fit a large amount of data. On the other hand, PSM-based methods \cite{amin2013multi,kostrikov2014depth,PavlakosZDD17,belagiannis20143D,qiu2019cross} are more robust to occlusion and get better out-of-distribution generalization performance. It is worth noting that PSM 
is mainly used in the multiview setting. Our experiment in the monocular setting shows that PSM alone gets very bad results because of its limited capability to reduce ambiguity ($3$D pose estimates may still be inaccurate although their limb lengths are correct). The observation motivates us to combine PSM and GNN in order to benefit from their advantages. Note that the task is non-trivial because PSM requires solving the discrete optimization function.

\vspace{0.2em}

\begin{table}[ht]
    \centering
    \resizebox{3.3in}{!}{
    \begin{tabular}{c|c|L|M|M|M} 
    \thickhline
    Method & Formula & Voxel Based & End-to-End & Cyclic Graph & Limb Length Prior \\
    \hline  
    PSM \cite{amin2013multi, qiu2019cross}& \ref{eq:psm} & \textcolor{mygreen}{\cmark} &   \textcolor{red}{\xmark}    & \textcolor{red}{\xmark} & \textcolor{mygreen}{\cmark} \\
    GNN \cite{zhao2019semantic,Ci_2019_ICCV} & \ref{eq:lcn} &  \textcolor{red}{\xmark} & \textcolor{mygreen}{\cmark} & \textcolor{mygreen}{\cmark} & \textcolor{red}{\xmark} \\
    \rowcolor{mygray}
    \textbf{ContextPose (Ours)} & \ref{eq:contextpose_general} & \textcolor{mygreen}{\cmark} & \textcolor{mygreen}{\cmark} & \textcolor{mygreen}{\cmark} & \textcolor{mygreen}{\cmark} \\
    \thickhline
    \end{tabular}}
    \caption{Comparison of different context modeling methods. Please refer to Section \ref{sec:comparison} for more details.}
    \label{tab:comparison}
\end{table}

To that end, we present an approach termed as \emph{ContextPose} on top of the general formula which is inspired by the attention mechanism \cite{vaswani2017attention}. It is built on the voxel representation \cite{Iskakov_2019_ICCV,tu2020voxelpose} and allows enforcing soft limb length constraints by \emph{paying more attention to information passed between locations that satisfy limb length constraints}. More importantly, the approach avoids solving the discrete optimization problem and can be trained end-to-end. Table \ref{tab:comparison} briefly summarizes different methods.

\subsection{Overview}
\begin{figure*}
	\centering
	\includegraphics[width=6.9in]{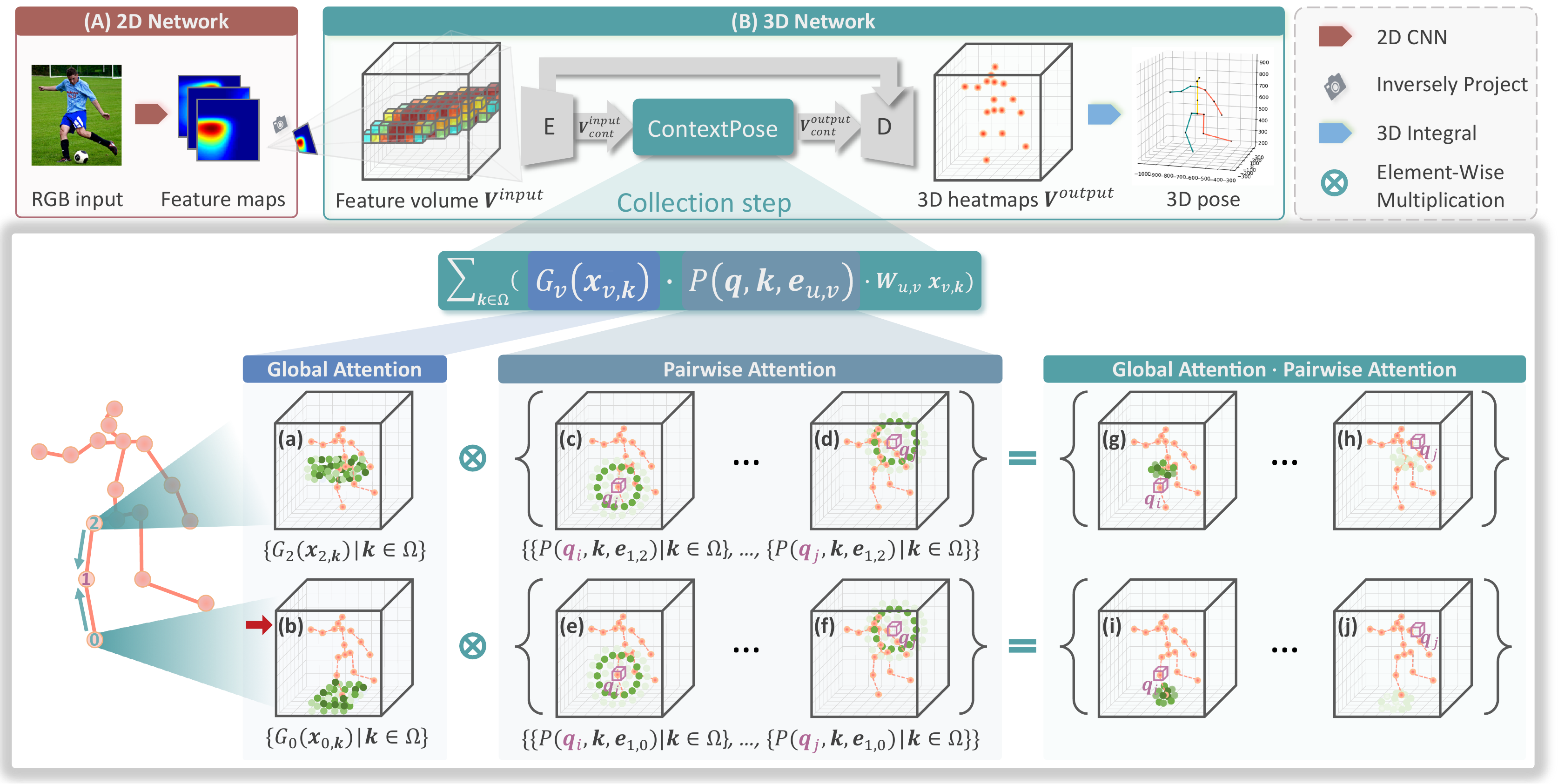}
	\caption{An example pipeline of using ContextPose for $3$D pose estimation. The bottom shows how ContextPose collects features from contextual joints based on global and pairwise attention.  Global attention, for example $G_0$ in \textbf{(b)}, represents the likelihood of $J_0$ at each voxel $\bm{k}$. For each voxel $\bm{q}$ of joint $J_1$, for example $\bm{q}_i$ in \textbf{(e)} or $\bm{q}_j$ in \textbf{(f)}, pairwise attention $P(\bm{q}, \bm{k}, \bm{e}_{1, 0})$ traverses every voxel $\bm{k}$ of joint $J_0$ and computes a spatial compatibility score between $\bm{q}$ and $\bm{k}$. The product of global attention and pairwise attention gives the weight in linear combination as shown in \textbf{(i)}-\textbf{(j)}.}
	\label{fig:overview}
\end{figure*}

Figure \ref{fig:overview} shows how ContextPose is leveraged by the state-of-the-art method \cite{Iskakov_2019_ICCV} for $3$D pose estimation. Given an input image, it first estimates $2$D features by a $2$D network (CNN). Then it inversely projects them to the $3$D voxels using camera parameters and uses a $3$D network to estimate $3$D heatmaps representing the likelihood of each voxel having each body joint. ContextPose can be inserted into the $3$D network to fuse features from different joints at different locations. Specifically, \emph{it updates the features of a joint at a voxel by a linear combination of the features of its contextual joints at all voxels}. The weights in linear combination are determined by their spatial relation (pairwise attention) and appearance (global attention) of the contextual joints. The bottom section of Figure \ref{fig:overview} shows more details of how we compute global attention and pairwise attention with the knee joint as an example.

In summary, we make three contributions:

\begin{itemize}
\item [1)] We develop a general formula for context modeling methods in $3$D human pose estimation which allows us to clearly understand their pros and cons. We also empirically compare them in a rigorous way.

\item [2)] We propose \emph{ContextPose} on top of the general formula which combines the advantages of PSM and GNN. In particular, it allows leveraging limb length constraints and can be leveraged by $3$D pose estimation networks for end-to-end training.

\item [3)] We demonstrate the state-of-the-art performance on two benchmark datasets. More importantly, ContextPose shows better generalization results on out-of-distribution data. The code and models will be released in order to inspire more research in this direction.
\end{itemize}

\section{Context Modeling: A Unified Perspective }
\label{sec:general_study}

\label{sec:intro}

We first introduce some notations and then reformulate PSM and GNN, respectively. Based on the reformulation, we develop a general formula for context modeling and show that both PSM and GNN are its special cases.

\subsection{Notations}

As shown in Figure \ref{fig:general_formula}, we represent human body by a graph $\mathcal{G} = (\mathcal{J}, \mathcal{E})$ where $\mathcal{J} = \{J_{0}, J_{1}, \dotsb, J_{N-1}\}$ represents $N$ body joints. The set $\mathcal{E}$ represents edges that connect pairs of joints. We define the joints that are connected by edges to be \textbf{contextual joints} of each other. The goal of monocular $3$D pose estimation is to estimate the $3$D  locations of the joints from a single image.

\subsection{Reformulate PSM}

PSM is commonly used in multiview $3$D pose estimation \cite{PavlakosZDD17,qiu2019cross}. It first divides the $3$D space by regular voxels $\Omega$ with each having a discrete location $\bm{q} \in \mathcal{R}^{3}$. The goal of PSM is to assign each joint to one of the voxels by minimizing an energy function defined on all joints. When the human graph is acyclic, PSM can be optimized by dynamic programming in which messages are sequentially passed from child nodes. In particular, the likelihood of a sub-tree with root joint $J_u$ at voxel $\bm{q}$ is computed as

\begin{equation}
\begin{small}
\label{eq:psm}
\begin{aligned}
    y_{u,\bm{q}}  = x_{u,\bm{q}} \cdot \prod_{J_v \in \text{child}(J_u)} (\max_{\bm{k} \in \Omega} \{ \psi(\bm{q},\bm{k}, \bm{e}_{u,v}) \cdot y_{v,\bm{k}} \}),
\end{aligned}
\end{small}
\end{equation}
where $\text{child}(J_u)$ denotes the children of $J_u$ and $x_{u,\bm{q}}$ is the confidence of $J_u$ at $\bm{q}$ determined by appearance. 

The formula can be interpreted by three steps: (1) for each non-leaf node $J_u$, it first collects features from each of its children $J_v$ by $\psi(\bm{q},\bm{k}, \bm{e}_{u,v}) \cdot  y_{v,\bm{k}}$ where $y_{v,\bm{k}}$ represents $J_v$'s likelihood of being at $\bm{k} \in \mathcal{R}^{3}$ which in turn is determined by its own children. The pairwise term $\psi(\bm{q},\bm{k}, \bm{e}_{u,v})$ encodes the limb length constraint measuring whether the distance between $\bm{q}$ and $\bm{k}$ satisfies the limb length prior in $\bm{e}_{u,v}$. The maximum score over all voxel locations $\Omega$ represents the message passed from joint $J_v$ to $J_u$. This step collects such information from all of its children; (2) then the context features collected from its children are aggregated by $\prod$; (3) finally, it updates $y_{u,\bm{q}}$ by multiplying the aggregated context with the confidence $x_{u,\bm{q}}$.

\subsection{Reformulate GNN}
Ci \etal \cite{Ci_2019_ICCV} present a formula which unifies FCN \cite{martinez2017simple}, GNN \cite{zhao2019semantic}, and LCN \cite{Ci_2019_ICCV}. We further reformulate it such that it has a similar form as PSM
\begin{equation}
\label{eq:lcn}
\begin{aligned}
    \bm{y}_u  = f(\bm{x}_u, \sum_{J_v \in \mathcal{J}} (\bm{e}_{u,v} \cdot \bm{W}_{u, v} \bm{x}_v)),
\end{aligned}
\end{equation}
where $\bm{x}_u \in \mathcal{R}^{M_{input}}$ represents the features of $J_u$ obtained from the previous layer or input, and $\bm{y}_u \in \mathcal{R}^{M_{output}}$ denotes the updated features of $J_u$. It is important to note that these methods do not discretize the $3$D space but directly estimate continuous locations. So we do not compute features for each discrete location $\bm{q}$ as in Eq. (\ref{eq:psm}). The binary scalar $\bm{e}_{u,v}$ encodes the pairwise relation between joint $J_u$ and $J_v$, and is set to be one if $J_v$ is a contextual joint of $J_u$. $\bm{W}_{u, v} \in \mathcal{R}^{M_{output} \times M_{input}}$ is a learnable weight matrix. We can also interpret the formula by three steps in a similar way as PSM. It first collects features by $\bm{e}_{u,v} \cdot \bm{W}_{u, v} \bm{x}_v$ from its contextual joints, then aggregates them using the sum operator $\sum$ and finally uses multilayer perceptron (MLP) $f$ to update the joint of interest.

The difference between FCN, GNN and LCN lies in how to compute $\bm{e}_{u,v}$ and $\bm{W}_{u, v}$. \emph{FCN} \cite{martinez2017simple} does not use human graph when collecting features. Instead, $\bm{e}_{u,v}$ is set to be one for every joint pair $(J_u, J_v)$. In contrast, in \emph{GNN} \cite{zhao2019semantic} and \emph{LCN} \cite{Ci_2019_ICCV}, $\bm{e}_{u,v}$ is set with special consideration. Generally, $\bm{e}_{u,v}$ is non-zero only when the two joints are connected according to the human graph. In other words, they only collect features from contextual joints. So their main difference lies in the collection step. Please refer to \cite{Ci_2019_ICCV} for more details.

\subsection{General Formula}
We introduce a general context modeling formula, which updates features $\bm{y}_{u}$ of joint $J_u$ by
\begin{equation}
\label{eq:general_formulaion}
\begin{aligned}
    \bm{y}_{u}  = f(\bm{x}_{u}, \: \text{AGG}(\{\: \phi(\bm{x}_{v}, \bm{e}_{u, v}) \: | \: \forall (J_u, J_v) \in \mathcal{E} \})\:),
\end{aligned}
\end{equation}
where $\bm{x}_u$ denotes the features of joint $J_u$ before updating, and $\bm{e}_{u, v}$ encodes the spatial relation prior (\eg limb length) between $J_u$ and $J_v$. There are three steps in the formula as will be detailed in the following.

\paragraph{1. Collection} For each joint of interest, it collects features from its contextual joints as represented by $\phi(\cdot, \cdot)$ in the formula. This is the most complex step in context modeling which determines where and how to collect features from the graph nodes.  
\vspace{-0.5em}

\paragraph{2. Aggregation} This is denoted by $\text{AGG}(\cdot)$ in the formula. It is a permutation invariant function, \eg sum or product function, defined on a set of contextual features. It aims to aggregate the collected features.
\vspace{-0.5em}

\paragraph{3. Update} This is denoted by $f(\cdot,\cdot)$ in the formula. It updates the feature of a joint by transforming its own as well as the aggregated features. 
\vspace{0.5em}

It is straightforward to verify that both PSM and GNN can be interpreted by the formula. The advantage of PSM is that it can explicitly enforce limb length constraints while GNN can learn implicit priors from a large amount of data. In the following, we present an approach to combine their advantages on top of the general formula.   

\section{ContextPose}
\label{sec:ContextPose}
This section introduces the details of \emph{ContextPose}. We first present an overview of how it can be leveraged by an existing method \cite{Iskakov_2019_ICCV} to estimate $3$D human pose in Section \ref{sec:arch_overview}. Then we dive into the technical and training details of ContextPose in the following three sub-sections. Finally, we discuss the differences between ContextPose and other context modeling methods in Section \ref{sec:comparison}.

\subsection{Architecture Overview}
\label{sec:arch_overview}

We adopt the state-of-the-art $3$D pose estimator \cite{Iskakov_2019_ICCV} as our baseline. As shown in Figure \ref{fig:overview}, it first constructs a $3$D feature volume by inversely projecting image features to the $3$D space using camera parameters. Then the feature volume is fed to an encoder-decoder network to estimate $3$D heatmaps. In particular, it predicts $N$ scores for each voxel representing the likelihood of $N$ joints. Finally, we compute expectation over the $3$D heatmaps of each joint to obtain its $3$D location \cite{sun2018integral}. ContextPose is inserted between the encoder and decoder network.

\subsection{ContextPose}
\label{sec:detail_ContextPose}
Denote the input tensor of ContextPose as $\bm{V}^{input}_{cont} \in \mathcal{R}^{NM  \times D \times H \times W}$ which represents the features of $N$ joints at $D \times H \times W$ voxels. 
We split $\bm{V}^{input}_{cont}$ into $N$ groups along the channel dimension such that each group corresponds to the features of one joint. Inspired by the attention mechanism \cite{vaswani2017attention}, ContextPose updates the features of a joint $J_u$ at voxel $\bm{q}$ by a linear combination of the features of its contextual joints at all voxels
\begin{equation}
\begin{small}
\label{eq:contextpose_general}
\begin{aligned}
    \bm{y}_{u,\bm{q}} = \bm{x}_{u,\bm{q}} + \sum_{J_v \in \mathcal{J}}[\sum_{\bm{k} \in \Omega}( G_v(\bm{x}_{v,\bm{k}}) \cdot P(\bm{q},\bm{k}, \bm{e}_{u,v}) \cdot \bm{W}_{u, v} \bm{x}_{v,\bm{k}})],
\end{aligned}
\end{small}
\end{equation}
where $\Omega$ denotes the set of voxels, $\bm{x}_{v,\bm{k}} \in \mathcal{R}^{M}$ denotes the features of joint $J_v$ at voxel $\bm{k}$. The global attention $G_v(\bm{x}_{v,\bm{k}})$ and pairwise attention $P(\bm{q},\bm{k}, \bm{e}_{u,v})$ determines the weight in linear combination. $\bm{W}_{u, v} \in \mathcal{R}^{M \times M} $ is a learnable matrix to transform features. \\

\noindent\textbf{Global Attention (GA)}
We estimate a confidence score for each joint $J_v$ at a voxel $\bm{k}$ representing to what extent should this feature contribute to other joints. Intuitively, we expect a lower score for non-person voxels in order to reduce the risk of corrupting good features. In other words, we expect large scores for voxels that are likely to include joint $J_v$. As a result, joint $J_u$ can focus on features from high likelihood voxels of joint $J_v$ (see Figure \ref{fig:overview} \textbf{(a)} and \textbf{(b)}). The GA for joint $J_v$ is defined as
\begin{equation}
\begin{aligned}
    G_v(\bm{x}_{v,\bm{k}}) \propto \emph{exp}(\bm{d}^{T}_{v}\bm{x}_{v,\bm{k}}),
\end{aligned}
\label{eq:global}
\end{equation}
which is normalized such that $\sum_{\bm{k} \in \Omega} G_v(\bm{x}_{v,\bm{k}}) = 1$. $\bm{d}_{v} \in \mathcal{R}^{M}$ is a learnable vector. \\

\noindent\textbf{Pairwise Attention (PA)} PA explores spatial relation between a pair of joints. The general idea is to give larger weights to features passed from locations of a joint that satisfy the pre-defined spatial relation. In this work, we focus on limb length constraints. But this can be extended to other priors such as limb orientations. If joint $J_v$ is connected to $J_u$ by a rigid bone, then their distance in the $3$D space is fixed for the same person which is independent of human postures. Offline, we compute the average distance $\mu_{u,v}$ and the standard deviation $\sigma_{u,v}$ in the training set as the limb length distribution prior and let $\bm{e}_{u,v} = (\mu_{u,v}, \sigma_{u, v})$ as the limb pre-defined parameters. The pairwise attention for the joint pair is defined as
\begin{equation}
\begin{aligned}
    P(\bm{q},\bm{k}, \bm{e}_{u,v}) \propto \emph{exp}(- \frac{ (||\bm{q}-\bm{k}||_2-\mu_{u,v})^2}{2\alpha\sigma_{u,v}^2 + \epsilon}).
\end{aligned}
\label{eq:pairwise}
\end{equation}
The pairwise attention is normalized over all voxels such that $\sum_{\bm{k} \in \Omega} G_v(\bm{x}_{v,\bm{k}}) \cdot P(\bm{q},\bm{k}, \bm{e}_{u,v}) = 1$. The hyper-parameter $\alpha$ is used to adjust the tolerance to limb length errors, which is empirically set to be $1500$ in this work. The parameter $\epsilon$ is used to improve numerical robustness. See Figure \ref{fig:overview} {{\textbf{(c)}-\textbf{(f)}}}. Besides, if joint $J_v$ is not connected to $J_u$ by a rigid bone, the features from joint $J_v$ may also be helpful to $J_u$. For example, left hand may also help the detection of right hand. In this case, we simply set the pairwise term to be $P(\bm{q},\bm{k}, \bm{e}_{u,v})=1$ and completely rely on the global attention to determine the weights.

\subsection{Regression of 3D Human Pose}
\label{sec:losses}
The decoder network transforms the output of ContextPose $\bm{V}^{output}_{cont} \in \mathcal{R}^{NM \times D \times H \times W}$ to $3$D heatmaps $\bm{V}^{output} \in \mathcal{R}^{N \times D\textprime \times H\textprime \times W\textprime}$ of $N$ body joints which represents the likelihood of each joint at each location. Then the $3$D location $\bm{J}_u$ for joint $J_u$ is obtained by computing the expectation of $\bm{V}^{output}_u \in \mathcal{R}^{D\textprime \times H\textprime \times W\textprime}$ with the common integral technique \cite{sun2018integral} according to the following formula
\begin{equation}
\label{eq:loss}
\begin{aligned}
    \bm{J}_u = \sum_{x=1}^{D\textprime}\sum_{y=1}^{H\textprime}\sum_{z=1}^{W\textprime}(x, y, z) \cdot \bm{V}_{u}^{output}(x, y, z).
\end{aligned}
\end{equation}

\subsection{Training}
The parameters in ContextPose are jointly learned with the $2$D CNN and the encoder-decoder network by enforcing two losses:
\begin{equation}
\label{eq:loss}
\begin{aligned}
    \mathcal{L} = \mathcal{L}_{3D} + \lambda \mathcal{L}_{GA},
\end{aligned}
\end{equation}
in which $\mathcal{L}_{3D}$ and $\mathcal{L}_{GA}$ are the loss functions enforced on the $3$D joint locations and global attention maps, respectively.

Same as \cite{Iskakov_2019_ICCV}, we compute the $L_1$ loss between the ground-truth $3$D pose $\bm{J}^{gt}$ and the estimated $3$D pose $\bm{J}$ with a weak heatmap regularizer which promotes Gaussian shape distribution for the estimated $3$D heatmaps as
\begin{equation}
\label{eq:3D loss}
\begin{aligned}
    \mathcal{L}_{3D} = \frac{1}{N} \sum_{J_u \in \mathcal{J}} (||\bm{J}_{u} - \bm{J}_{u}^{gt}||_1 - \beta \cdot log(\bm{V}_{u}^{output}(\bm{J}_{u}^{gt}))).
\end{aligned}
\end{equation}

In addition, to help the GA focus on the voxels that are likely to have joint $J_u$, we enforce an $L_2$ loss:

\begin{equation}
\label{eq:GA loss}
\begin{aligned}
    \mathcal{L}_{GA} = \frac{1}{NDHW} \sum_{J_u \in \mathcal{J}} ||\bm{G}_u - \bm{G}_u^{gt}||^2_2,
\end{aligned}
\end{equation}
where $\bm{G}_u \in \mathcal{R}^{D \times H \times W}$ is the GA map for joint $J_u$ and $\bm{G}_u^{gt} \in \mathcal{R}^{D \times H \times W}$ is the ground-truth heatmap generated by applying a $3$D Gaussian centered at the ground truth location of the joint $J_u$. 

In our experiment, we set $\beta$ and $\lambda$ to be $10^{-2}$ and $10^6$.

\subsection{Comparison of PSM, GNN and ContextPose}
\label{sec:comparison}

It is easy to verify that PSM, GNN, and ContextPose are all special cases of the general formula Eq. (\ref{eq:general_formulaion}). The main difference between them lies in the \emph{collection} step which includes the structures of human graph $\mathcal{G}$, pairwise relation $\bm{e}_{u,v}$, the collection function $\phi(\cdot, \cdot)$, and training scheme. We will compare them side by side from the above aspects hoping to clearly understand their advantages and disadvantages. \\

\noindent\textbf{Graph Structures} PSM often uses acyclic graphs in order to get optimum solution. In contrast, ContextPose is not subject to this restriction. Cyclic graph offers greater flexibility to represent more powerful and natural context. For example, in ContextPose, we can add connections between left and right shoulders to the human graph and require that they cannot be at the same location which helps solve the ``double counting'' problem. We can even add connections between joints in neighboring frames to promote smoothness in future work. GNN can also use cyclic graphs but it cannot explicitly express and enforce natural rules on the joints. It is not clear what kind of pairwise relation does GNN learns from data which makes it a black box.   \\  

\noindent\textbf{Pairwise Relation} In PSM, the pairwise relation is often implemented as limb length constraints. As discussed in Eq. (\ref{eq:psm}), it encourages detections of a pair of joints that satisfy the limb length prior. In GNN, the pairwise term reflects the similarity between the features of two nodes. Although the features also encode some location information, it is hardly possible that GNN will implicitly learn limb length constraints. ContextPose does not enforce hard limb length constraints as PSM. But it encourages pose estimates to have reasonable limb length by focusing on features that are passed between locations that satisfy limb length constraints. \\

\noindent\textbf{End-to-End Learning}
PSM requires solving a discrete optimization problem in order to obtain optimal locations for all joints. In particular, it uses the argmax operator to identify optimal voxels for each joint which makes the approach non-differentiable. In contrast, the GNN-based methods can be trained end-to-end because all operators in the collection, aggregation and update functions are differentiable. ContextPose can also be trained end-to-end which combines the advantages of PSM and GNN. \\

\noindent\textbf{Quantization Error}
The PSM-based methods and ContextPose both work on discrete voxels. So their accuracy depends on the size of each voxel. Using a smaller voxel decreases quantization error but meanwhile increases computation time. In \cite{Iskakov_2019_ICCV}, the authors propose to compute expectation over the heatmaps to obtain continuous $3$D locations which notably decreases the impact of quantization. 

\section{Experiments}
\label{sec:experiments}
\subsection{Datasets}
\noindent\textbf{Human3.6M (H36M)\cite{h36m_pami}} Following \cite{Ci_2019_ICCV}, we use the subjects S$1$, S$5$, S$6$, S$7$, and S$8$ for training, and S$9$, S$11$ for testing. The Mean Per Joint Position Error (MPJPE) metric is computed under two protocols: Protocol \#1 computes MPJPE between the ground-truth (GT) and the estimated $3$D poses after aligning their root (mid-hip) joints; Protocol \#2 reports MPJPE after the $3$D estimate is aligned with the GT via a rigid transformation. Additionally, we present two new metrics to comprehensively measure the quality of the $3$D pose estimates: (1) Mean Per Limb Length Error (MPLLE) computes the average limb length error between the GT and estimated poses over $16$ limbs (\ie the purple edges in Figure \ref{fig:general_formula}), and (2) Mean Per Limb Angle Error (MPLAE) measures the average limb angle error between the GT and the estimated poses.\\

\noindent\textbf{MPI-INF-3DHP (3DHP) \cite{mehta2017monocular}} This dataset provides monocular videos of six subjects acting in three different scenes which include green screen indoor scenes, indoor scenes and outdoor scenes. This dataset is often used to evaluate the generalization performance of different models. Following the convention, we directly apply our model trained on the H36M dataset to this dataset without re-training. We report results using two metrics: Percentage of Correctly estimated Keypoints (PCK) \cite{andriluka14cvpr} and Area Under the Curve (AUC) \cite{mehta2017monocular}. 

\subsection{Implementation Details}
\label{sec:implementation}
We use the state-of-the-art $3$D pose estimator \cite{Iskakov_2019_ICCV} as our baseline to estimate $3$D poses. We insert ContextPose between the encoder and decoder networks as shown in Figure \ref{fig:overview}. To reduce GPU memory cost, we decrease the number of layers in the $3$D network from five to two. The modification slightly improves the results of the baseline. For the ContextPose network, $M$ is set to be $3$. We jointly train the $2$D and $3$D networks for $30$ epochs with the Adam \cite{kingma2015adam} optimizer. 
The learning rates are set to be  $0.0001$ and $0.001$ for the $2$D and $3$D networks, respectively. To prevent from over-fitting to the human appearance in the H36M dataset, we fix the $2$D network and train the $3$D network for $20$ epochs before end-to-end training.

\begin{table*}[]
\center
\small
\setlength{\tabcolsep}{2pt}
\resizebox{6.7in}{!}{
\begin{tabular}{l c c c c c c c c c c c c c c c c}
\thickhline 
\textbf{Protocol \#1} & Dire. & Disc. & Eat & Greet & Phone & Photo & Pose & Purch. & Sit & SitD & Smoke & Wait & WalkD & Walk & WalkT & Avg \\
\hline 

Zhou \etal \cite{zhou2017towards} ICCV'17 & 54.8 & 60.7 & 58.2 & 71.4 & 62.0 & 65.5 & 53.8 & 55.6 & 75.2 & 111.6 & 64.2 & 66.1 & 51.4 & 63.2 & 55.3 & 64.9 \\
Martinez \etal (\textbf{FCN}) \cite{martinez2017simple} ICCV'17 & 51.8 & 56.2 & 58.1 & 59.0 & 69.5 & 78.4 & 55.2 & 58.1 & 74.0 & 94.6 & 62.3 & 59.1 & 65.1 & 49.5 & 52.4 & 62.9 \\
Pavlakos \etal \cite{pavlakos2018ordinal} CVPR'18 & 48.5 & 54.4 & 54.4 & 52.0 & 59.4 & 65.3 & 49.9 & 52.9 & 65.8 & 71.1 & 56.6 & 52.9 & 60.9 & 44.7 & 47.8 & 56.2 \\
Yang \etal \cite{Yang_2018_CVPR} CVPR'18 & 51.5 & 58.9 & 50.4 & 57.0 & 62.1 & 65.4 & 49.8 & 52.7 & 69.2 & 85.2 & 57.4 & 58.4 & 43.6 & 60.1 & 47.7 & 58.6 \\
Zhao \etal (\textbf{GNN}) \cite{zhao2019semantic} CVPR'19 & 47.3 & 60.7 & 51.4 & 60.5 & 61.1 & 49.9 & 47.3 & 68.1 & 86.2 & \textbf{55.0} & 67.8 & 61.0 & 42.1 & 60.6 & 45.3 & 57.6 \\
Qiu \etal (\textbf{PSM}) \cite{qiu2019cross} ICCV'19 & 223.1 & 231.8 & 273.0 & 237.3 & 248.1 & 243.9  & 209.0 & 279.7 & 280.9 & 296.3 & 241.9 & 234.0 & 230.8 & 217.8 & 220.4 & 244.8\\ 
Iskakov \etal \cite{Iskakov_2019_ICCV} ICCV'19 & 41.9 & 49.2 & 46.9 & 47.6 & 50.7 & 57.9 & 41.2 & 50.9 & 57.3 & 74.9 & 48.6 & 44.3 & \textbf{41.3} & 52.8 & 42.7 & 49.9 \\
Wang \etal \cite{wang2019not} ICCV'19 & 44.7 & 48.9 & 47.0 & 49.0 & 56.4 & 67.7 & 48.7 & 47.0 & 63.0 & 78.1 & 51.1 & 50.1 & 54.5 & 40.1 & 43.0 & 52.6 \\
Ci \etal (\textbf{LCN}) \cite{Ci_2019_ICCV} ICCV'19 & 46.8 & 52.3 & 44.7 & 50.4 & 52.9 & 68.9 & 49.6 & 46.4 & 60.2 & 78.9 & 51.2 & 50.0 & 54.8 & 40.4 & 43.3 & 52.7 \\
Pavllo* \etal \cite{pavllo:videopose3d:2019} CVPR'19 & 47.1 & 50.6 & 49.0 & 51.8 & 53.6 & 61.4 & 49.4 & 47.4 & 59.3 & 67.4 & 52.4 & 49.5 & 55.3 & 39.5 & 42.7 & 51.8* \\
Cai* \etal \cite{cai2019exploiting} ICCV'19 & 46.5 & 48.8 & 47.6 & 50.9 & 52.9 & 61.3 & 48.3 & 45.8 & 59.2 & 64.4 & 51.2 & 48.4 & 53.5 & 39.2 & 41.2 & 50.6* \\
Xu* \etal \cite{Xu_2020_CVPR} CVPR'20 & 40.6 & 47.1 & 45.7 & 46.6 & 50.7 & 63.1 & 45.0 & 47.7 & 56.3 & 63.9 & 49.4 & 46.5 & 51.9 & 38.1 & 42.3 & \underline{49.2}* \\
\rowcolor{mygray}

Ours & \textbf{36.3} & \textbf{42.8} & \textbf{39.5} & \textbf{40.0} & \textbf{43.9} & \textbf{48.8} & \textbf{36.7} & \textbf{44.0} & \textbf{51.0} & 63.1 & \textbf{44.3} & \textbf{40.6} & 44.4 & \textbf{34.9} & \textbf{36.7} & \textbf{43.4} \\


\thickhline
\textbf{Protocol \#2} & Dire. & Disc. & Eat & Greet & Phone & Photo & Pose & Purch. & Sit & SitD & Smoke & Wait & WalkD & Walk & WalkT & Avg \\
\hline 

Martinez \etal (\textbf{FCN}) \cite{martinez2017simple} ICCV'17 & 39.5 & 43.2 & 46.4 & 47.0 & 51.0 & 56.0 & 41.4 & 40.6 & 56.5 & 69.4 & 49.2 & 45.0 & 49.5 & 38.0 & 43.1 & 47.7 \\
Pavlakos \etal \cite{pavlakos2018ordinal} CVPR'18 & 34.7 & 39.8 & 41.8 & 38.6 & 42.5 & 47.5 & 38.0 & 36.6 & 50.7 & 56.8 & 42.6 & 39.6 & 43.9 & 32.1 & 36.5 & 41.8 \\
Yang \etal \cite{Yang_2018_CVPR} CVPR'18 & \textbf{26.9} & \textbf{30.9} & 36.3 & 39.9 & 43.9 & 47.4 & 28.8 & \textbf{29.4} & \textbf{36.9} & 58.4 & 41.5 & \textbf{30.5} & \textbf{29.5} & 42.5 & 32.2 & \underline{37.7} \\
Qiu \etal (\textbf{PSM}) \cite{qiu2019cross} ICCV'19 & 117.0 & 123.2 & 128.0 & 121.7 & 126.1 & 128.7 & 105.3 & 130.1 & 145.1 & 170.2 & 125.1 & 114.5 & 128.9 & 115.3 & 117.1 & 126.7 \\
Wang \etal \cite{wang2019not} ICCV'19 & 33.6 & 38.1 & 37.6 & 38.5 & 43.4 & 48.8 & 36.0 & 35.7 & 51.1 & 63.1 & 41.0 & 38.6 & 40.9 & 30.3 & 34.1 & 40.7 \\
Ci \etal (\textbf{LCN}) \cite{Ci_2019_ICCV} ICCV'19 & 36.9 & 41.6 & 38.0 & 41.0 & 41.9 & 51.1 & 38.2 & 37.6 & 49.1 & 62.1 & 43.1 & 39.9 & 43.5 & 32.2 & 37.0 & 42.2 \\
Pavllo* \etal \cite{pavllo:videopose3d:2019} CVPR'19 & 36.0 & 38.7 & 38.0 & 41.7 & 40.1 & 45.9 & 37.1 & 35.4 & 46.8 & 53.4 & 41.4 & 36.9 & 43.1 & 30.3 & 34.8 & 40.0* \\
Cai* \etal \cite{cai2019exploiting} ICCV'19 & 36.8 & 38.7 & 38.2 & 41.7 & 40.7 & 46.8 & 37.9 & 35.6 & 47.6 & \textbf{51.7} & 41.3 & 36.8 & 42.7 & 31.0 & 34.7 & 40.2* \\
Xu* \etal \cite{Xu_2020_CVPR} CVPR'20 & 33.6 & 37.4 & 37.0 & 37.6 & 39.2 & 46.4 & 34.3 & 35.4 & 45.1 & 52.1 & 40.1 & 35.5 & 42.1 & 29.8 & 35.3 & 38.9* \\
\rowcolor{mygray}
Ours & 30.5 & 34.9 & \textbf{32.0} & \textbf{32.2} & \textbf{35.0} & \textbf{37.8} & \textbf{28.6} & 32.6 & 40.8 & 52.0 & \textbf{35.0} & 31.9 & 35.6 & \textbf{26.6} & \textbf{28.5} & \textbf{34.6} \\

\thickhline 
\end{tabular}}
\caption{The MPJPE (mm) of the state-of-the-art methods on the H36M dataset under protocol \#1 and protocol \#2, respectively. \emph{*} means the method uses temporal information in videos.}
\label{tab:state_of_the_art_h36m}
\end{table*}

\subsection{Comparison to the State-of-the-arts}
\begin{figure}
	\centering
	\includegraphics[width=3.1in]{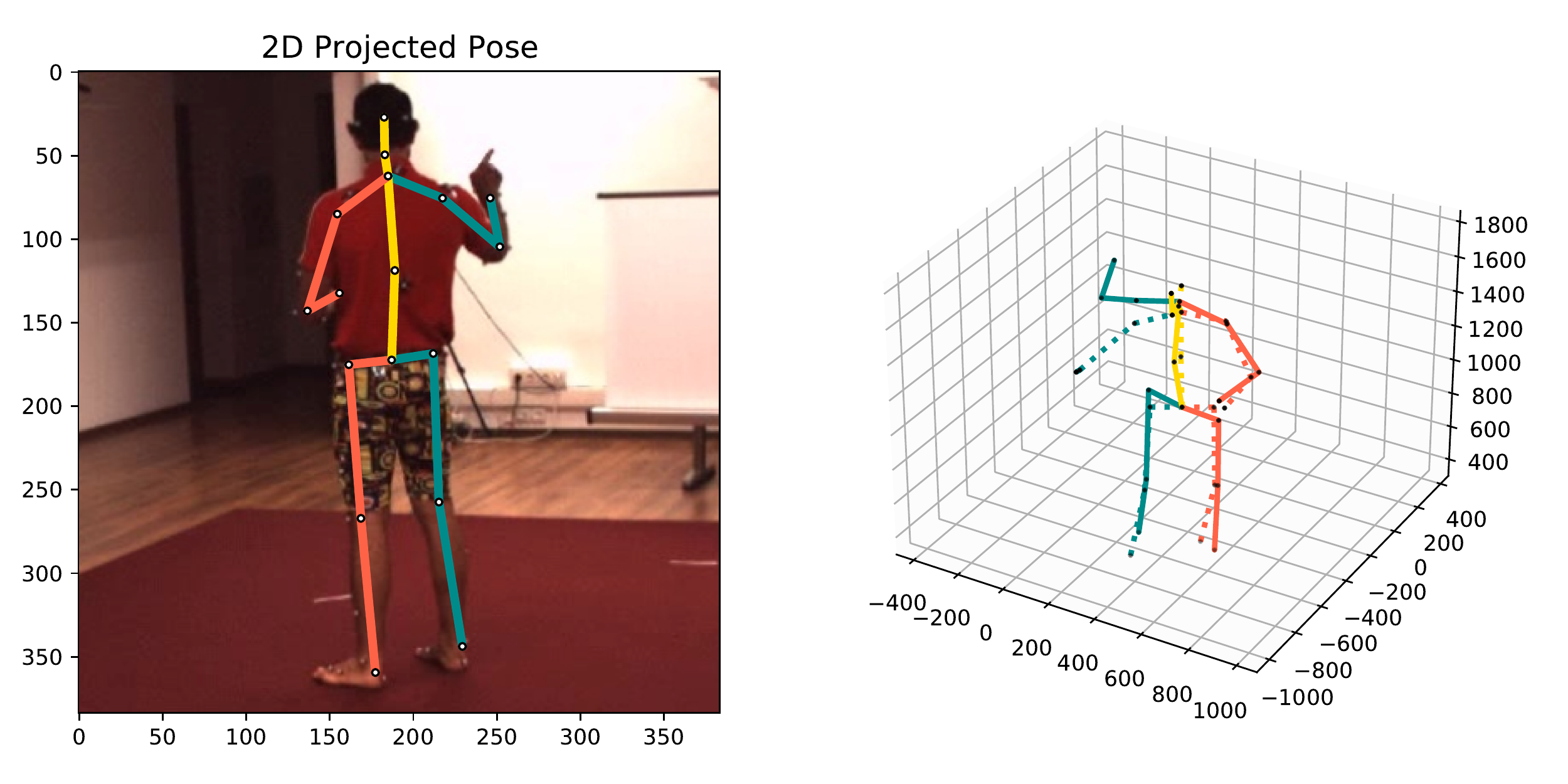}
	\caption{Visualization of a $3$D pose estimated by PSM \cite{qiu2019cross}. The left figure shows the projection of the estimated $3$D pose. The right figure shows the estimated (solid lines) and GT (dashed lines) $3$D poses. The estimated $3$D pose has correct $2$D projection but it is very different from GT $3$D pose. It means PSM suffers from severe ambiguity when it is used in the monocular setting.}
	\label{fig:PSM_result}
\end{figure}

\noindent\textbf{Results on the H36M Dataset} Table \ref{tab:state_of_the_art_h36m} shows the results of the state-of-the-art methods on the H36M dataset. Our approach outperforms the state-of-the-art methods by a notable margin under both protocols. This includes methods that explore temporal information in videos (labeled by * in the table). In particular, our method outperforms PSM \cite{qiu2019cross}, FCN \cite{martinez2017simple}, GNN \cite{zhao2019semantic}, and LCN \cite{Ci_2019_ICCV} by an even larger margin which validates the effectiveness of our context modeling strategy. We discover in our experiment that PSM \cite{qiu2019cross} gets very poor results in the monocular setting. To investigate the reasons, we project the estimated $3$D poses back to $2$D images and find that, for most cases, the projections perfectly match the $2$D people although their $3$D estimates are very different from the GT poses. We show an example in Figure \ref{fig:PSM_result}. This is mainly because PSM alone has limited capability to resolve  ambiguity. Note that a $3$D pose estimate may be inaccurate even when its limb lengths are correct. In contrast, the deep learning-based methods such as GNN \cite{martinez2017simple,zhao2019semantic,Ci_2019_ICCV} have strong capability to reduce ambiguity because they can fit a large amount of data. We will discuss in more details on why our approach gets more accurate estimates than PSM and GNN in the subsequent ablative study.\\

\noindent\textbf{Results on the 3DHP Dataset}
Table \ref{tab:state_of_the_art_3dhp} shows the results of different methods on the 3DHP dataset. Our approach achieves significantly better PCK and AUC scores than other methods including FCN, LCN, and PSM for almost all scenes. The result suggests that ContextPose has strong generalization performance which we think is due to the leverage of limb length priors in deep networks. FCN \cite{martinez2017simple} gets a low accuracy because the dense connections degrade the generalization capability which has already been discussed in \cite{Ci_2019_ICCV}. LCN \cite{Ci_2019_ICCV} gets better results by fusing features of contextual joints but it is still worse than ours. The result validates the importance of combining deep networks and limb length priors.

\begin{table}
    \centering
    \resizebox{3.3in}{!}{
    \begin{tabular}{l|SMLMM} 
    \thickhline 
    Method & GS (PCK) & noGS (PCK) & Outdoor (PCK) & ALL (PCK) $\uparrow$ & ALL (AUC) $\uparrow$ \\
    \hline
    \multicolumn{6}{c}{Trained on: H36M+MPII \cite{andriluka14cvpr}} \\
    \hline
    \rowcolor{mygray}
    Zhou \etal \cite{zhou2017towards} & 71.1 & 64.7 & 72.7 & 69.2 & 32.5 \\
    \rowcolor{mylightergray}
    Yang \etal \cite{Yang_2018_CVPR} & - & - & - & 69.0 & 32.0 \\
    \rowcolor{mygray}
    Wang \etal \cite{wang2019not} & - & - & - & 71.9 & 35.8 \\
    \hline
    \multicolumn{6}{c}{Trained on: H36M+MPII+LSP \cite{Johnson10}} \\
    \hline
    \rowcolor{mygray}
    Pavlakos \etal \cite{pavlakos2018ordinal} & 76.5 & 63.1 & 77.5 & 71.9 & 35.3 \\
    \hline
    \multicolumn{6}{c}{Trained on: H36M} \\
    \hline
    \rowcolor{mygray}
    Martinez \etal (\textbf{FCN}) \cite{martinez2017simple} & 49.8 & 42.5 & 31.2 & 42.5 & 17.0 \\
    \rowcolor{mylightergray}
    Qiu \etal (\textbf{PSM}) \cite{qiu2019cross} & 26.4 & 22.6 & 19.6 & 23.3 & 8.0 \\
    \rowcolor{mygray}
    Ci \etal (\textbf{LCN}) \cite{Ci_2019_ICCV} & 74.8 & 70.8 & \textbf{77.3} & \underline{74.0} & \underline{36.7} \\
    \rowcolor{mylightergray}
    Baseline & 75.2	& 73.3 & 62.2 & 71.3 & 35.0 \\
    \hline
    \rowcolor{mygray}
    Ours & \textbf{82.6} & \textbf{80.5} & \textbf{77.3} & \textbf{80.5} & \textbf{42.7} \\
    \thickhline
    \end{tabular}}
    \caption{The results of the state-of-the-art methods on the 3DHP dataset. GS represents the green screen background scene. The results of \cite{martinez2017simple} are taken from \cite{luo2018orinet}.}
    \label{tab:state_of_the_art_3dhp}
\end{table}

\subsection{Ablation Study}

\begin{table}
    \centering
    \resizebox{3.3in}{!}{
    \begin{tabular}{l|c|c|c|c|c|c|c|c} 
    \thickhline 
    \multirow{2}{*}{Method} & \multirow{2}{*}{GA} & \multirow{2}{*}{PA} & \multicolumn{3}{c|}{S9} &  \multicolumn{3}{c}{S11} \\
    \cline{4-9}
    & & & MPJPE $\downarrow$ & MPLLE $\downarrow$ & MPLAE $\downarrow$ & MPJPE $\downarrow$ & MPLLE $\downarrow$ & MPLAE $\downarrow$ \\
    \hline
    Baseline & \textcolor{red}{\xmark} & \textcolor{red}{\xmark} & 54.38 & 15.03 & 0.1600 & 35.12 & 10.16 & 0.1250  \\	
    Ours w/o PA & \textcolor{mygreen}{\cmark} & \textcolor{red}{\xmark} & 52.00 & 14.58 & 0.1517 & 35.16 & 9.78 & 0.1240 \\
    Ours w/o GA & \textcolor{red}{\xmark} & \textcolor{mygreen}{\cmark} &  52.46 & 14.16 & 0.1524 & 34.98 & 9.67 & 0.1224 \\
    \cellcolor{mygray}{Ours} & \cellcolor{mygray}{\textcolor{mygreen}{\cmark}} & \cellcolor{mygray}{\textcolor{mygreen}{\cmark}} & \cellcolor{mygray}{\textbf{50.24}} & \cellcolor{mygray}{\textbf{14.13}} & \cellcolor{mygray}{\textbf{0.1509}} & \cellcolor{mygray}{\textbf{34.10}} & \cellcolor{mygray}{\textbf{9.50}} & \cellcolor{mygray}{\textbf{0.1217}} \\
    \thickhline
    \end{tabular}}
    \caption{Ablative study on the global attention and pairwise attention in ContextPose. We show the MPJPE (mm), MPLLE (mm) and MPLAE (radian) on each test subject separately. ContextPose achieves large improvement on the more challenging subject of S$9$.}
    \label{tab:ablation_effect}
\end{table}

\noindent\textbf{Effect of ContextPose}
\label{sec:ablation_effect}
We first compare our approach to the baseline w/o ContextPose. The results on the H36M dataset are shown in Table \ref{tab:ablation_effect}. We can see that ContextPose notably decreases MPJPE of the baseline from $54.38$mm to $50.24$mm on the challenging subject S$9$. MPLLE decreases by nearly $6\%$ meaning that the limb lengths of the estimated $3$D poses are more accurate than the baseline. The improvement for S$11$ in terms of MPJPE is marginal because the baseline is already very accurate. However, we can see that there is still clear improvement in terms of limb lengths and angles. The result of the baseline is different from the number in Table \ref{tab:state_of_the_art_h36m} because we use a smaller $3$D network in Table \ref{tab:ablation_effect} to reduce memory usage as stated in Section \ref{sec:implementation}. 

\begin{figure}
	\centering
	\includegraphics[width=3.1in]{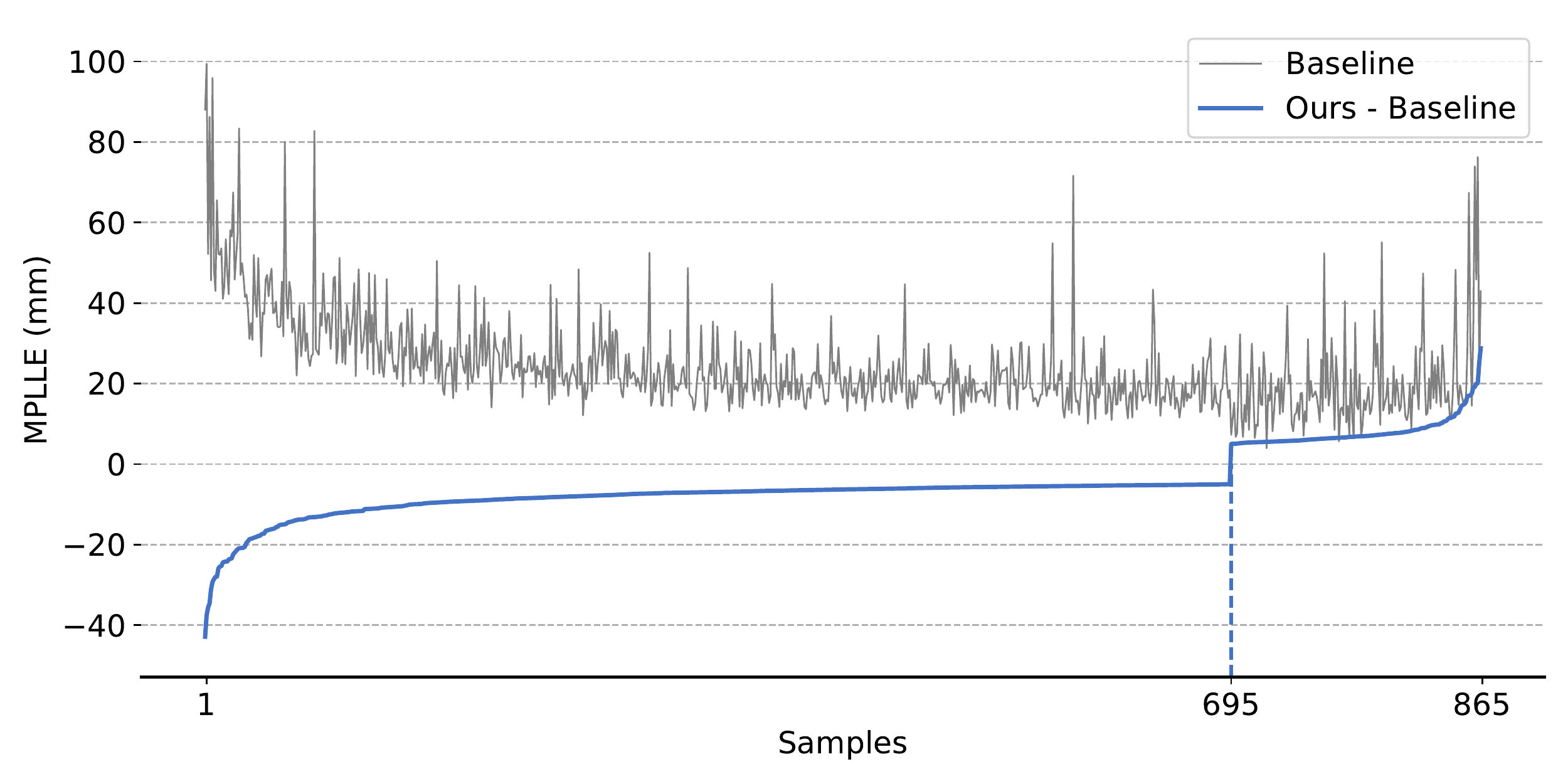}
	\caption{MPLLE (mm) of individual samples. The gray line shows the errors of the baseline. The blue line represents the error difference between ContextPose and baseline (below zero means our method gets smaller error).}
	\label{fig:delta_error}
\end{figure}

We plot the MPLLE of the baseline and our method for each sample in H36M dataset in Figure \ref{fig:delta_error}. We can see that ContextPose gets smaller errors than baseline for about $80\%$ of the test data. In particular, the improvement is larger for hard cases where the baseline gets large errors (see the left side of the figure). It indicates that ContextPose reduces the chance of getting absurd poses by exploring context. There are few cases where ContextPose gets worse results. This usually happens when multiple body joints are occluded which makes estimating global attention a very challenging task.

Table \ref{tab:state_of_the_art_3dhp} shows the results on the 3DHP dataset. We can see that using ContextPose significantly improves the PCK of the baseline from $71.3\%$ to $80.5\%$. The result represents that ContextPose is very important to improve the generalization performance of the $3$D pose estimator. This is a big advantage for actual deployment. In fact, we can see that our approach even outperforms the methods which use even more training data. \\

\noindent\textbf{Effect of GA and PA}
We report results when we add one of the two modules (GA and PA) to the baseline in Table \ref{tab:ablation_effect}. Adding only the GA module makes little difference on the ultimate results measured by MPJPE, MPLLE, and MPLAE. In contrast, if we add the PA module, the results are improved by a notable margin which validates the importance of pairwise compatibility in context modeling. \\

\begin{figure}
	\centering
	\includegraphics[width=3.3in]{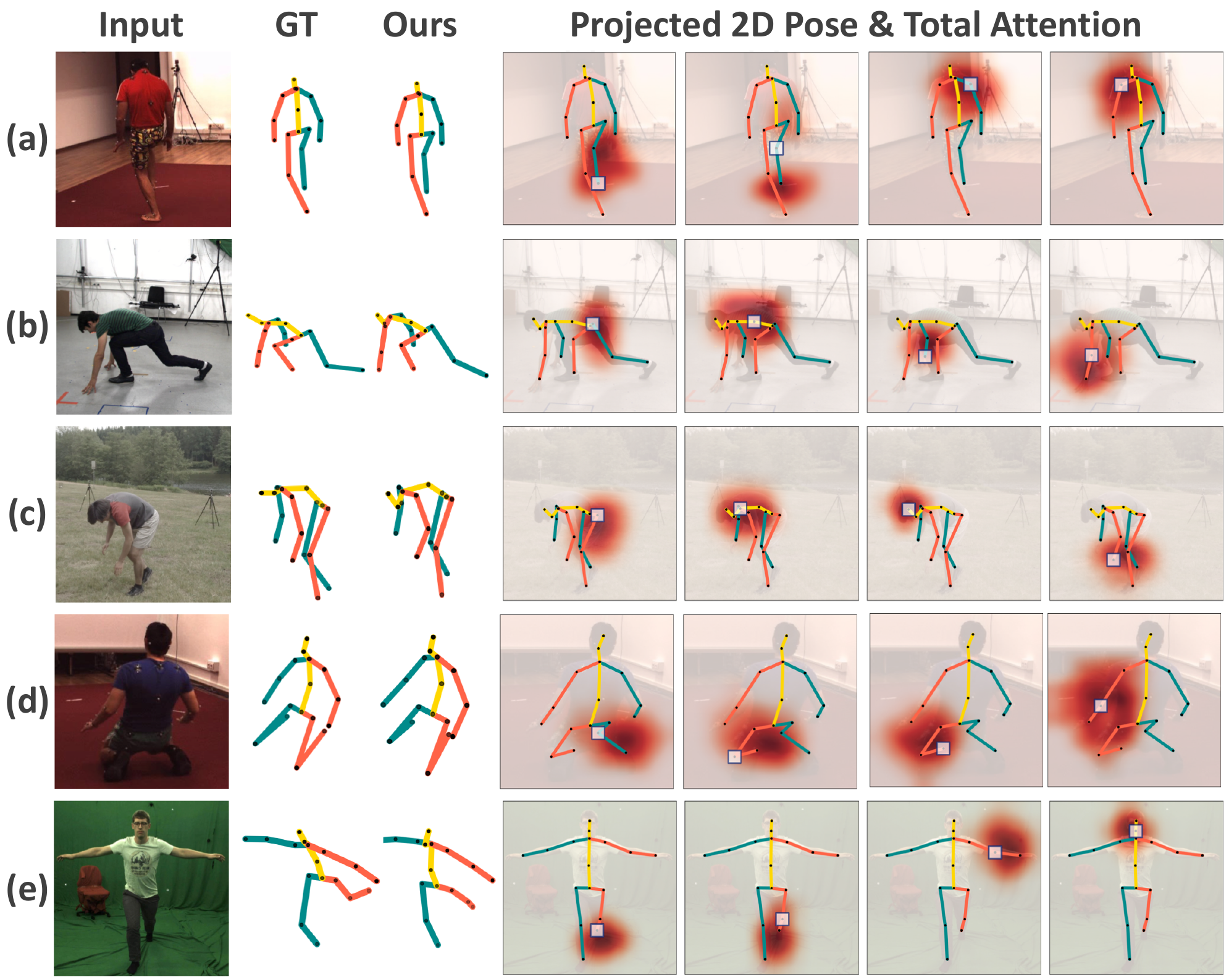}
	\caption{Example $3$D pose estimates. The last four columns show the projected $2$D poses and the weights in linear combination for some random joints (highlighted by small blue boxes). Row (d) and (e) show two failure cases.}
	\label{fig:quality_results}
\end{figure}

\subsection{Qualitative Results}
Figure \ref{fig:quality_results} shows some $3$D poses estimated by ContextPose. The last four columns show the predicted weights (\ie the product of the GA and PA) for some random joints. In the first case of (a), the approach pays more attention to the features around the right knee when estimating the right ankle. Similarly, in the third case of (b), it focuses on features from right elbow when estimating right wrist. We show two failure cases in row (d) and (e). In particular, in (e) our estimate has correct limb lengths but inaccurate limb angles for the left leg. In addition, the projection of the $3$D pose is also reasonable. This is a common error for monocular $3$D pose estimation because it has severe ambiguity.

\section{Conclusion}
\label{sec:conclusion}
We first introduce a general formula for context modeling in $3$D pose estimation which allows comparing PSM and GNN side by side. Based on the formula, we present ContextPose that combines their advantages which allows enforcing limb length constraints in deep networks. So it can be trained end-to-end on large data. The approach outperforms the state-of-the-art methods on two benchmarks, and more importantly, shows better generalization performance on unseen datasets. 

\section*{Acknowledgement}
This work was supported in part by National Key R\&D Program of China (2018YFB1403900), NSFC-61625201 and NSFC-62061136001.

{\small
\bibliographystyle{ieee_fullname}
\bibliography{egbib}
}


\end{document}